\documentclass[%
 reprint,
 amsmath,amssymb,
 aps,
 showkeys,
pra,
]{revtex4-2}

\usepackage{graphicx}% Include figure files
\usepackage{dcolumn}% Align table columns on decimal point
\usepackage{bm}% bold math
\usepackage{hyperref}% add hypertext capabilities
\usepackage{bbm}
\newtheorem{definition}{Definition}
\newtheorem{lemma}{Lemma}

\begin{document}

\preprint{APS/123-QED}

\title{Efficient Finite Initialization with Partial Norms for Tensorized Neural Networks and Tensor Networks Algorithms}% Force line breaks with \\

\author{Alejandro Mata Ali}
\email{alejandro.mata.ali@gmail.com}
\affiliation{Instituto Tecnológico de Castilla y León, Burgos, Spain}
\author{Iñigo Perez Delgado}
\email{iperezde@ayesa.com}
\affiliation{i3B Ibermatica Fundazioa, Parque Tecnológico de Bizkaia, Ibaizabal Bidea, Edif. 501-A, 48160 Derio, Spain}
\author{Marina Ristol Roura}
\email{mristol@ayesa.com}
\affiliation{i3B Ibermatica Fundazioa, Parque Tecnológico de Bizkaia, Ibaizabal Bidea, Edif. 501-A, 48160 Derio, Spain}
\author{Aitor Moreno Fdez. de Leceta}
\email{aitormoreno@lksnext.com}
\affiliation{Quantum Technologies and Systems Unit, LKS Next, MONDRAGON Corporation, Goiru 7, 20500 Arrasate-Mondragón, Gipuzkoa, Spain}

\date{\today}% It is always \today, today,

\begin{abstract}
We present two algorithms to initialize layers of tensorized neural networks and general tensor network algorithms using partial computations of their Frobenius norms and positive lineal entrywise sums, depending on the type of tensor network involved. The core of this method is the use of the norm of subnetworks of the tensor network in an iterative way, so that we normalize by the finite values of the norms that led to the divergence or zero norm. In addition, the method benefits from the reuse of intermediate calculations. We have also applied it to the Matrix Product State/Tensor Train (MPS/TT) and Matrix Product Operator/Tensor Train Matrix (MPO/TT-M) layers and have seen its scaling versus the number of nodes, bond dimension, and physical dimension. All code is publicly available.
\end{abstract}

%Use showkeys class option if keyword
\maketitle

%\tableofcontents

\section{Introduction}
\textit{Deep neural networks} \cite{DeepLearning} are widely used in machine learning to achieve good results in industrial, research, and various other applications. This good performance has led to its use in more complex contexts, requiring a larger number of parameters. Consequently, various architectures have been utilized to enhance its performance. The greatest example are \textit{Large Language Models} (LLM)~\cite{LLaMa}, which make use of an extreme number of parameters, requiring large devices to use them. The memory requirements are a huge limitation in future applications and the scalability of the current line of progress in artificial intelligence.

With the advent of quantum computing applied to various fields, interest in quantum information compression methods~\cite{Quantum_Compression} has increased due to the exponential capacity of quantum systems to handle information. This is one of the reasons for the \textit{Quantum Machine Learning} field. However, the current limitation of quantum hardware does not allow one to implement most of the desired algorithms and models that quantum computing offers.

\begin{figure}
    \centering
    \includegraphics[width=\linewidth]{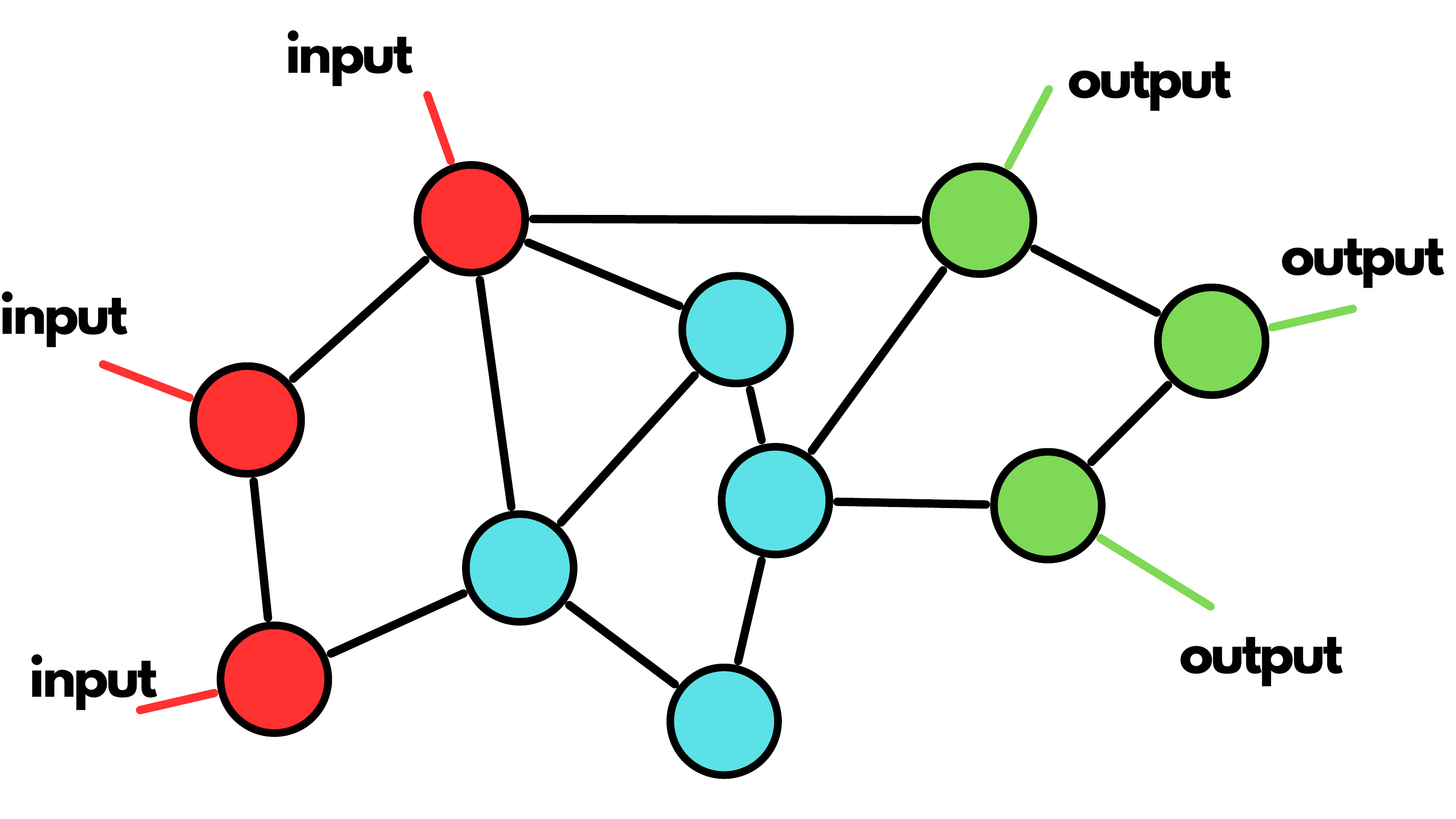}
    \caption{Arbitrary tensor network layer.}
    \label{fig:Arbit Layer}
\end{figure}

In this context, researchers have explored different classical techniques which could take advantage of the properties of quantum systems, the \textit{quantum-inspired}. This allows some of the advantages of quantum computing without the need of a quantum device to implement the operations. One of the best known quantum-inspired techniques is the \textit{Tensor networks}~\cite{Tensor,OrusTN}, which are graphical representations of tensor algebra calculations. Tensor networks have a great capacity to ``compress" tensor information by means of representations such as \textit{Matrix Product State/Tensor Trains} (MPS/TT) \cite{MPS1} or \textit{Projected Entangled-Pair States} (PEPS) \cite{MPS2}. This allows us to retain the most important information of the represented tensors with several fewer parameters. This compression has been applied to several machine learning models in various ways, such as decomposing matrices into tensor networks~\cite{Tensorizing,Compress_DNN_MPO}. It has been applied to \textit{neural networks}~\cite{NN_Compression}, \textit{convolutional neural networks}~\cite{convo_compression}, \textit{transformers}~\cite{Transformer_compress}, \textit{spiking neural networks}~\cite{snn_tn} or LLM~\cite{LLM_quantum_tn,Compactifai}. Tensor networks have also been used as the main model, directly training the tensor network itself \cite{LowRank,Convolutional} (Fig. \ref{fig:Arbit Layer}).

Our focus of analysis will be on methods in which a large tensor network is generated to model the layer and trained directly, rather than using a trained dense matrix and compressing it. For example, when we try to use \textit{tensorized physics-informed neural networks}~\cite{TPINN} or tensor neural networks to solve differential equations~\cite{TNNPhysics} for large industrial cases, such as the heat equation of an engine or fluids in a turbine. In this type of cases, an initialization problem is often encountered \cite{Gradient}, the explosion or vanishing of the represented tensor values. If we initialize the elements of each tensor of the tensor network with a certain distribution, when we contract the tensor network to obtain the tensor it represents, some of its final elements are too large (infinite) or too small (null) for the computer. This problem may also arise in some quantum machine learning algorithms in which the aim is to compress a classical machine learning model.

If we want to eliminate these problems, a first proposal could be to contract the tensor network and eliminate or reduce these elements. However, in very large layers we cannot store all the final tensor elements in memory, so it is not possible in this way. In addition, we may not know how to properly rescale those divergent elements without damaging the other ones. One way is to reinitialize the tensor network by changing a distribution with better hyperparameters, changing the mean and standard deviation, or rescaling them~\cite{Gradient}. However, many of these methodologies are not easy to apply in all cases in an efficient way.

In this work, we present two novel algorithms to address this problem in two different scenarios. The first is applicable to tensor networks of general values, which consists of iteratively calculating the \textit{Frobenius norm} for different sections of the tensor network until a condition is met, when we divide all the parameters of the tensor network by the factor calculated in a particular way. This allows us to gradually make the Frobenius norm of the layer tend to the number we want, without having to repeatedly re-initialize. We call this method the \textit{Frobenius Tensor Network Renormalization} (FTNR). The second method applies to tensor networks whose represented entries are nonnegative, where the process is analogous to that of the previous method but involves calculating the partial positive \textit{lineal entrywise sum} of reduced forms. We call this method the \textit{Lineal Tensor Network Renormalization} (LTNR).

These algorithms are remarkably interesting for hierarchical tree form layers, especially in the \textit{Tensor Train} (TT), \textit{Tensor Train Matrix} (TT-M), and \textit{Projected Entangled Pair States} (PEPS). This can also be used in other methods with tensor networks, such as combinatorial optimization, to determine hyperparameters, and it can be combined with other initialization methods.

The main contributions of this work are two efficient algorithms to initialize a tensorized artificial intelligence model, or other tensor network algorithms, without an explosion or vanishing of the parameters, for large layers. This work is structured as follows. First, in Sec.~\ref{sec: description} we will describe properly the main problem we want to solve, and we will introduce a brief background on existing algorithms to deal with this. Then, in Secs.~\ref{sec: general initial} and \ref{sec: positive initial}, we will introduce our new algorithms for general and positive scenarios, respectively. Additionally, in Sec.~\ref{sec: experiments} we will perform several experiments to investigate the limitations of the algorithms. Finally, in Sec.~\ref{sec:other} we will analyze other possible applications of these algorithms.

We created a Python function to run it on an arbitrary PyTorch layer, available in a Jupyter Notebook in the GitHub repository \href{https://github.com/DOKOS-TAYOS/Efficient_Initialization_Tensor_Networks}{https://github.com/DOKOS-TAYOS/Efficient\_Initialization\_Tensor\_Networks} and a Streamlit demo in \href{https://efficient-initialization-tensor-networks.streamlit.app/}{https://efficient-initialization-tensor-networks.streamlit.app/}

\section{Description of the problem and background}\label{sec: description}

\begin{figure}[h]
    \centering
    \includegraphics[width=\linewidth]{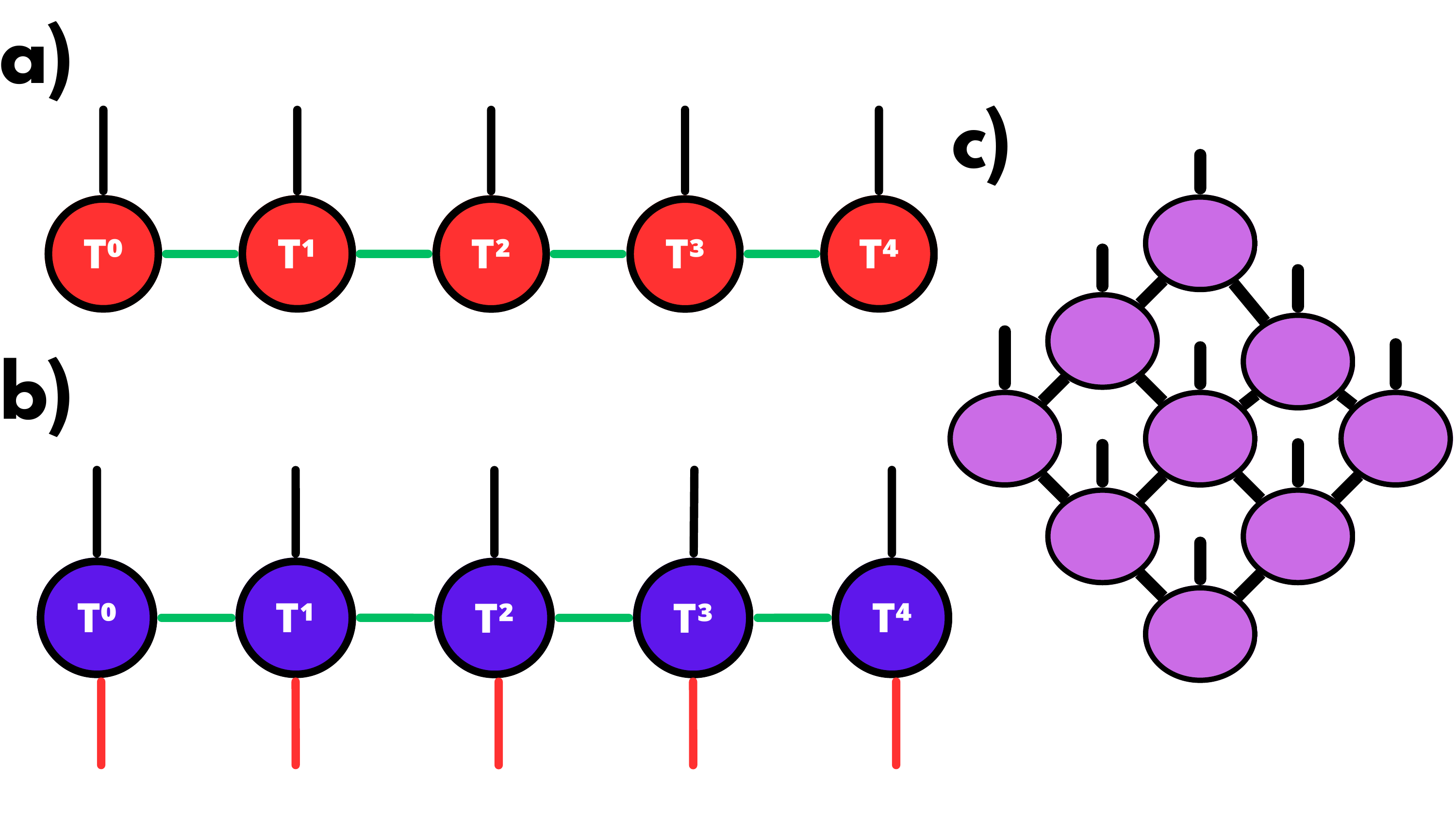}
    \caption{a) Tensor Train layer with 5 indexes. b) Tensor Train Matrix layer with 10 indexes. c) PEPS layer with 9 indexes.}
    \label{fig:TT_Layer}
\end{figure}

When we have a tensor network of $N$ nodes, we will find that the elements of the tensor representing the tensor network are given by the sum of a set of values, each given by the product of $N$ elements of the different nodes. If we examine the case of an TT layer, as in Fig. \ref{fig:TT_Layer}.a, the elements of the layer are given as
\begin{equation}
    T_{ijklm} = \sum_{n,o,p,q} T^0_{in} T^1_{njo} T^2_{okp} T^3_{plq} T^4_{qm}.
    \label{eq:TT_5}
\end{equation}

We see that for 5 indexes in the tensor we have to multiply 5 tensor elements, but in the general case with $N$ indexes we have
\begin{equation}
    T_{i_0i_1\dots i_{N-1}} = \sum_{j_0j_1\cdots j_{N-2}} T^0_{i_0j_0} T^1_{j_0i_1j_1}\dots T^{N-1}_{j_{N-2}i_{N-1}},
\end{equation}
multiplying $N$ elements of the tensors $T^i$ to obtain the element of the tensor $T$.

To exemplify the problem we want to solve, let us think about this. For a general case with bond dimension $b$, the dimension of the index that is contracted between each pair of nodes, $N$ nodes, and constant elements of the nodes with value $a$, we would have the following result for the represented tensor elements
\begin{equation}
    T_{i_0i_1\dots i_{N-1}} =  b^{N-1} a^N.
\end{equation}
We can see that with $N=20$ nodes, an element value of $a=1.5$ and a bond dimension of $b=10$, the final tensor elements would be $3.3\ 10^{22}$. This is a very large element for a good initialization. This is what we call the ``explosion" of the tensor values. On the other hand, if the value of the element was $a=0.01$, the tensor elements would be $10^{-20}$. This is the ``vanishing" of the tensor values. However, if we were to divide the values of these tensors by that number, we could arrive at the case where $a^N$ was a number too large or small for our computer to store, and we would get a 0 or infinite in all the renormalized elements.

This problem is exacerbated by the number of nodes in the layer, since each one is a product. Moreover, we cannot simply calculate these tensor elements for cases with many physical indexes, the output indexes. This is because the number of values to be kept in memory increases exponentially with the number of physical indexes. There is also the possibility that, instead of initializing the tensors with a similar distribution, we initialize them with different distributions, making some nodes in the network have smaller elements than other nodes in the network. However, this approach could lead us to problems with model training.

Tensor network architectures often entail a trade-off between expressivity and trainability, strongly influenced by the initial values of tensor cores. For conventional dense networks, random initialization~\cite{Glorot_Initial} is enough, making use of the \textit{Central Limit Theorem}. However, TN models require more complex strategies due to the multilinear nature of the tensor contractions and the role of bond dimensions, as we noted in the previous section.

The first approach is developed in~\cite{causal_TT_initial}, where they prepare the initialization of a TT for probabilistic models, as a ``warm-start". This technique allows to obtain a model which avoids the causal problems. However, it makes use of the TT-cross approximation~\cite{TT_Cross}, so it cannot be easily generalizable. The second approach is to generate unitary matrices from Haar measure distribution~\cite{Unitary}, because they preserve the norms. However, this has a limitation for general structures where unitarity is not enough, for example PEPS. The third approach is to connect some identity matrices with a small random noise term~\cite{tn4ml}. However, the authors indicate that this is a use-case specific technique.

\section{General Tensor network initialization protocol FTNR}\label{sec: general initial}
We first address the case of a general real value tensor network, where we have initialized each node with a similar distribution, allowing the elements of the nodes to be on the same scale with respect to each other. Our protocol is based on the use of partial Frobenius norms to normalize the total Frobenius norm of the resulting tensor.

\begin{figure}[h]
    \centering
    \includegraphics[width=\linewidth]{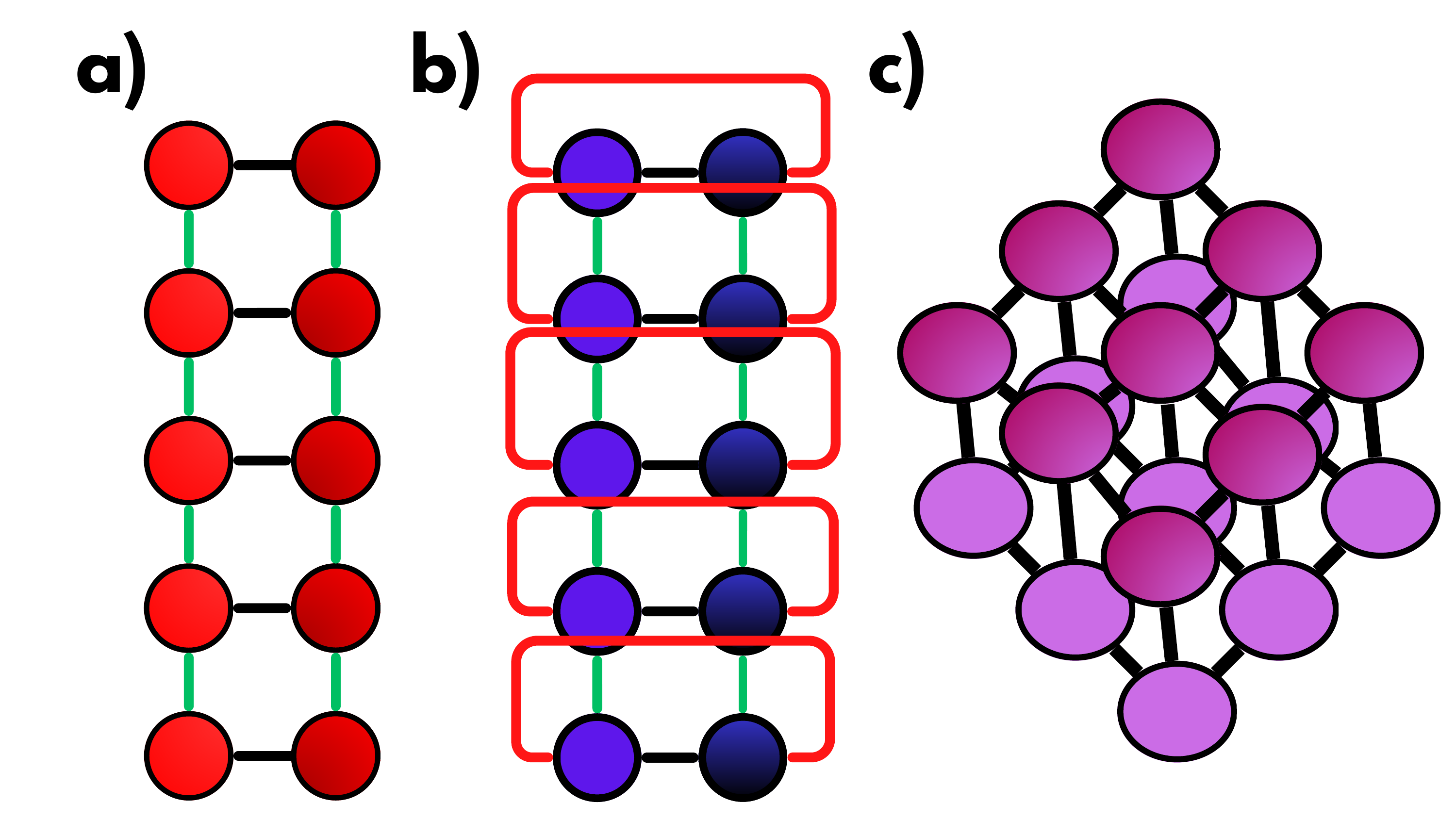}
    \caption{Square of the Frobenius norm calculated to a) Tensor Train layer. b) Tensor Train Matrix layer. c) PEPS layer.}
    \label{fig:Frob_tot}
\end{figure}

The \textit{Frobenius norm} of a matrix is given by the equation
\begin{equation}
    ||A||_F = \sqrt{\sum_{ij}|a_{ij}|^2} = \sqrt{\text{Tr} (A^\dagger A)}.
    \label{eq: Frobenius}
\end{equation}
In a tensor network, this corresponds to contracting the layer with a conjugate copy of itself so that each physical index is connected to the equivalent index of its copy. We can see some examples in Fig. \ref{fig:Frob_tot}.

The value of this closed contraction is $\langle A,A \rangle = ||A||_F^2$, not $||A||_F$ directly. The Frobenius norm itself is obtained only after taking the square root. In addition, this contraction can be computed without the need to calculate the elements of the represented matrix, using only the elements of the nodes.

The Frobenius norm is an indicator that serves to regularize layers of a model \cite{Anomaly} and gives an estimate of the order of magnitude of the matrix elements. With a smooth distribution of elements around $a_{00}$, the norm in Eq. \eqref{eq: Frobenius} will be of the order of $\sqrt{nm}\ |a_{00}|$ for a matrix $n\times m$.

To avoid the elements of the layer from being too large or too small and therefore having too large or too small outputs in initialization, we will normalize these elements so that the Frobenius norm of the tensor is a number that we choose, for example, $||A||_F=1$. This prevents our highest element from being higher than this value while taking advantage of the localized distribution of values to ensure that the smallest value is not too small.

Still, for a matrix $n \times m$ with entrywise scale of order one, a natural Frobenius target scales like $\sqrt{nm}$ rather than $nm$, since $||A||_F \approx \sqrt{nm}\ |a_{00}|$ under a smooth entrywise distribution.

For this purpose, we define what we call the partial square norm of the tensor network.

\subsection{Partial square norm of the tensor network}
For the remainder of the paper, we assume that we can consistently sort the nodes of a tensor network so that they form a single growing network.

\begin{definition}[Partial square norm]
$ $
\\
Given a tensor network $\mathcal{A}$ of $N$ nodes, and a tensor network $\mathcal{A}_n$ defined by the first $n$ nodes of $\mathcal{A}$, the partial square norm ${}^{pF}||\mathcal{A}||_{n,N}$ at the $n$ nodes of $\mathcal{A}$ is defined as the square of the Frobenius norm of $\mathcal{A}_n$. That is,
\begin{equation}
    {}^{pF}||\mathcal{A}||_{n,N} = ||\mathcal{A}_n||_F^2.
\end{equation}    
\end{definition}

\begin{lemma}[Multilinearity and scaling]
Let $A=\Phi(C_1,\ldots,C_N)$ be the contraction of a tensor network with $N$ cores. Then, for any scalars $\lambda_1,\ldots,\lambda_N$,
\begin{equation}
    \Phi(\lambda_1 C_1,\ldots,\lambda_N C_N) = \left(\prod_{k=1}^N \lambda_k\right)\Phi(C_1,\ldots,C_N).
\end{equation}
\end{lemma}

\noindent\textit{Proof.} Each entry of the contraction is a sum of products, and each product contains exactly one element from each core. Multiplying the $k$-th core by $\lambda_k$ multiplies every such product by $\lambda_k$. Factoring these scalars out of the sum gives the stated identity. \hfill$\square$

If we divide every core by $r>0$, then
\begin{equation}
    A' = r^{-N}A,
\end{equation}
and therefore every homogeneous norm of degree one satisfies
\begin{equation}
    ||A'|| = r^{-N}||A||.
\end{equation}
In particular,
\begin{equation}
    ||A'||_F^2 = r^{-2N}||A||_F^2.
\end{equation}
For a partial quantity built from the first $n$ cores, we likewise obtain
\begin{equation}
    S_n' = r^{-2n}S_n,
    \qquad
    L_n' = r^{-n}L_n,
\end{equation}
where $S_n=||\mathcal{A}_n||_F^2$ and, in the nonnegative case, $L_n=||\mathcal{A}_n||_L$.

\begin{figure}[h]
    \centering
    \includegraphics[width=\linewidth]{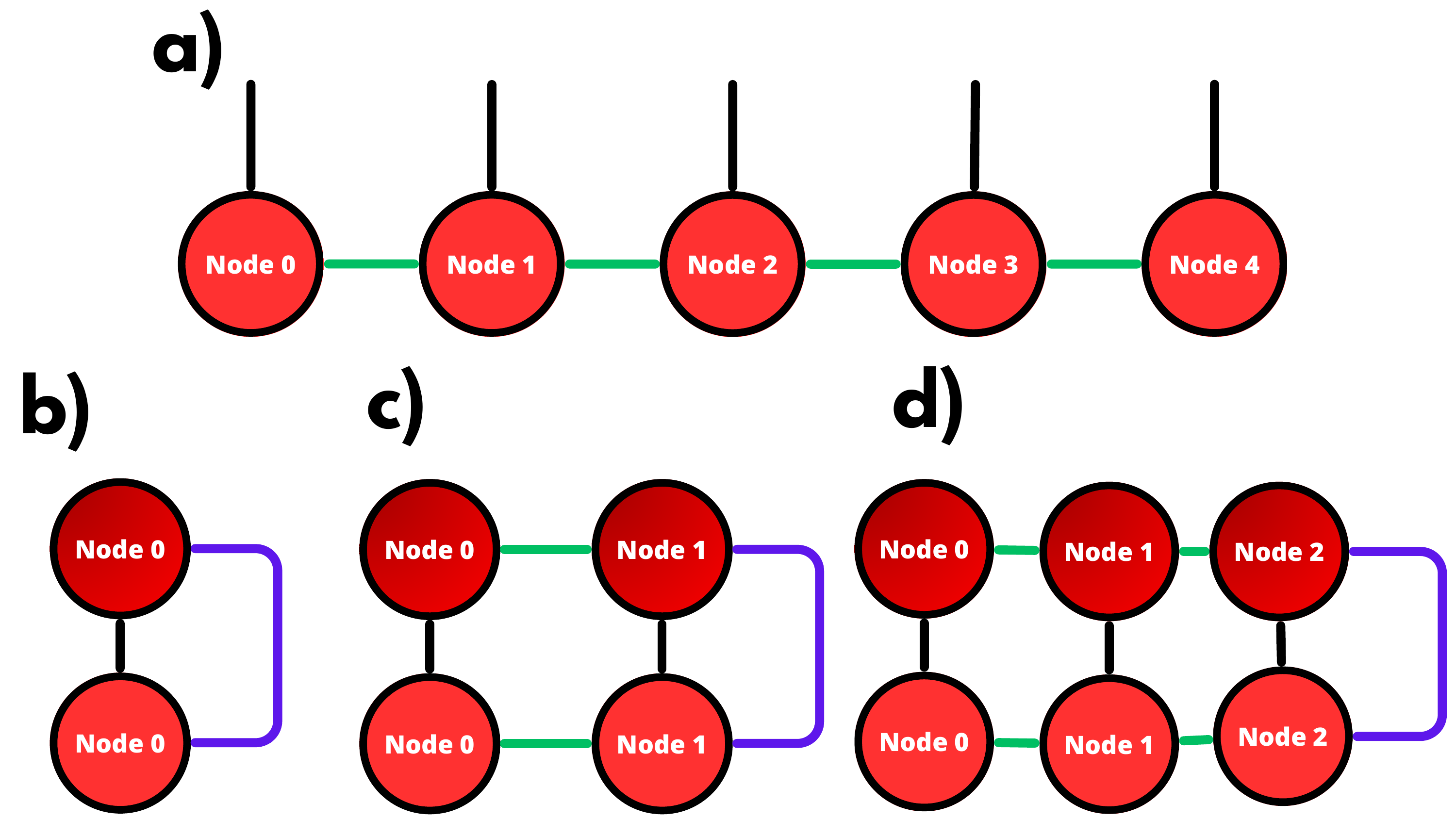}
    \caption{a) Tensor Train layer with 5 nodes. b) Partial square norm at 1 node. c) Partial square norm at 2 nodes. d) Partial square norm at 3 nodes.}
    \label{fig:TT_Partial}
\end{figure}

To get an idea of what this partial square norm is, we exemplify it with a simple case, a tensor train layer. We consider the tensor network in Fig. \ref{fig:TT_Partial}, whose nodes are sorted.

As we can see, in this case we would only have to do the same process as when calculating the total norm of the total tensor network, but stop at step $n$ and contract the bond index of the two final tensors of the chain.

We can see in the following Figs. \ref{fig:TT_Matrix} and \ref{fig:PEPS} how the partial square norm would be for a TT-Matrix layer and for a PEPS layer.
\begin{figure}[h]
    \centering
    \includegraphics[width=\linewidth]{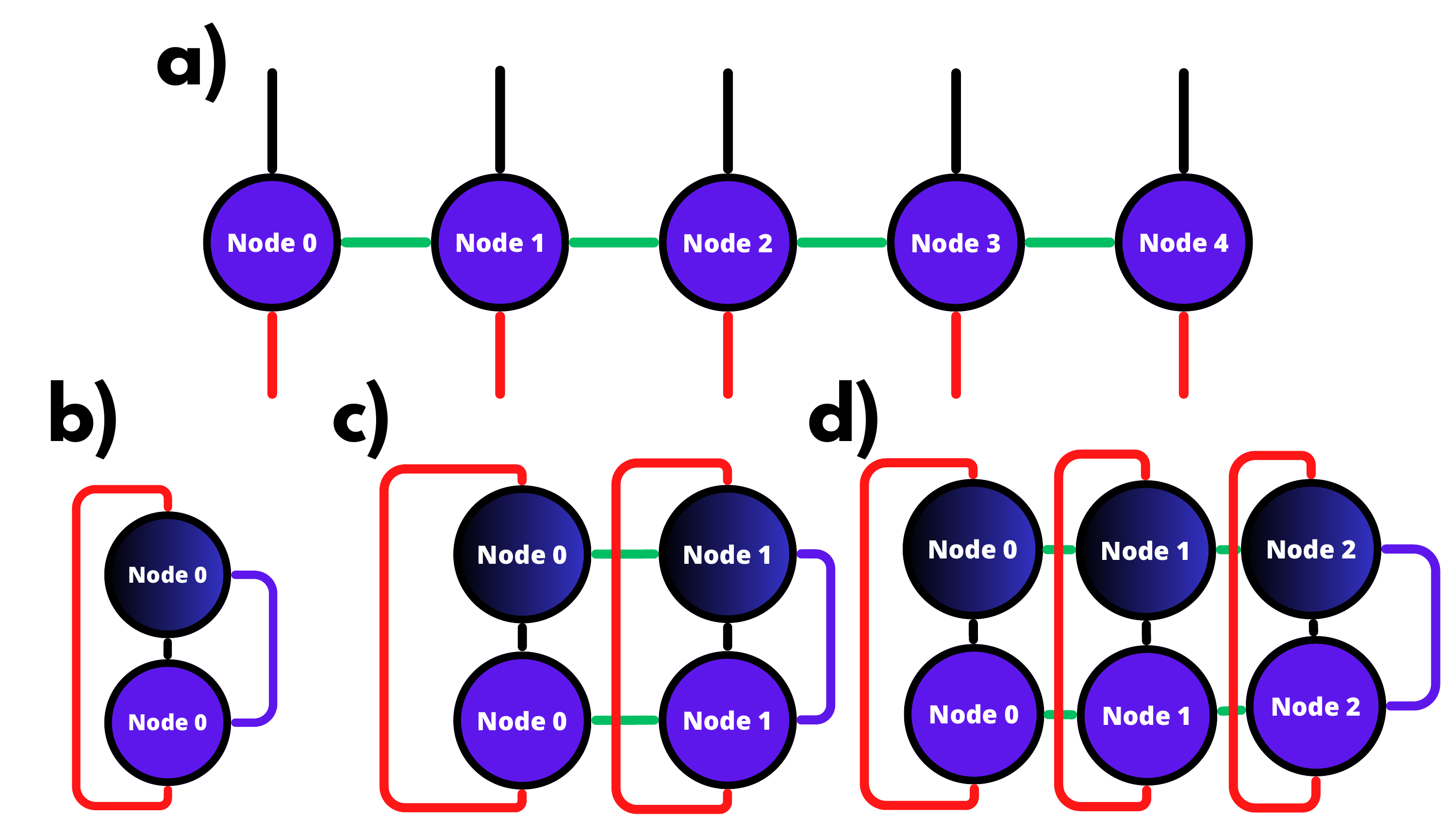}
    \caption{a) Tensor Train Matrix layer with 5 nodes. b) Partial square norm at 1 node. c) Partial square norm at 2 nodes. d) Partial square norm at 3 nodes.}
    \label{fig:TT_Matrix}
\end{figure}
\begin{figure}[h]
    \centering
    \includegraphics[width=\linewidth]{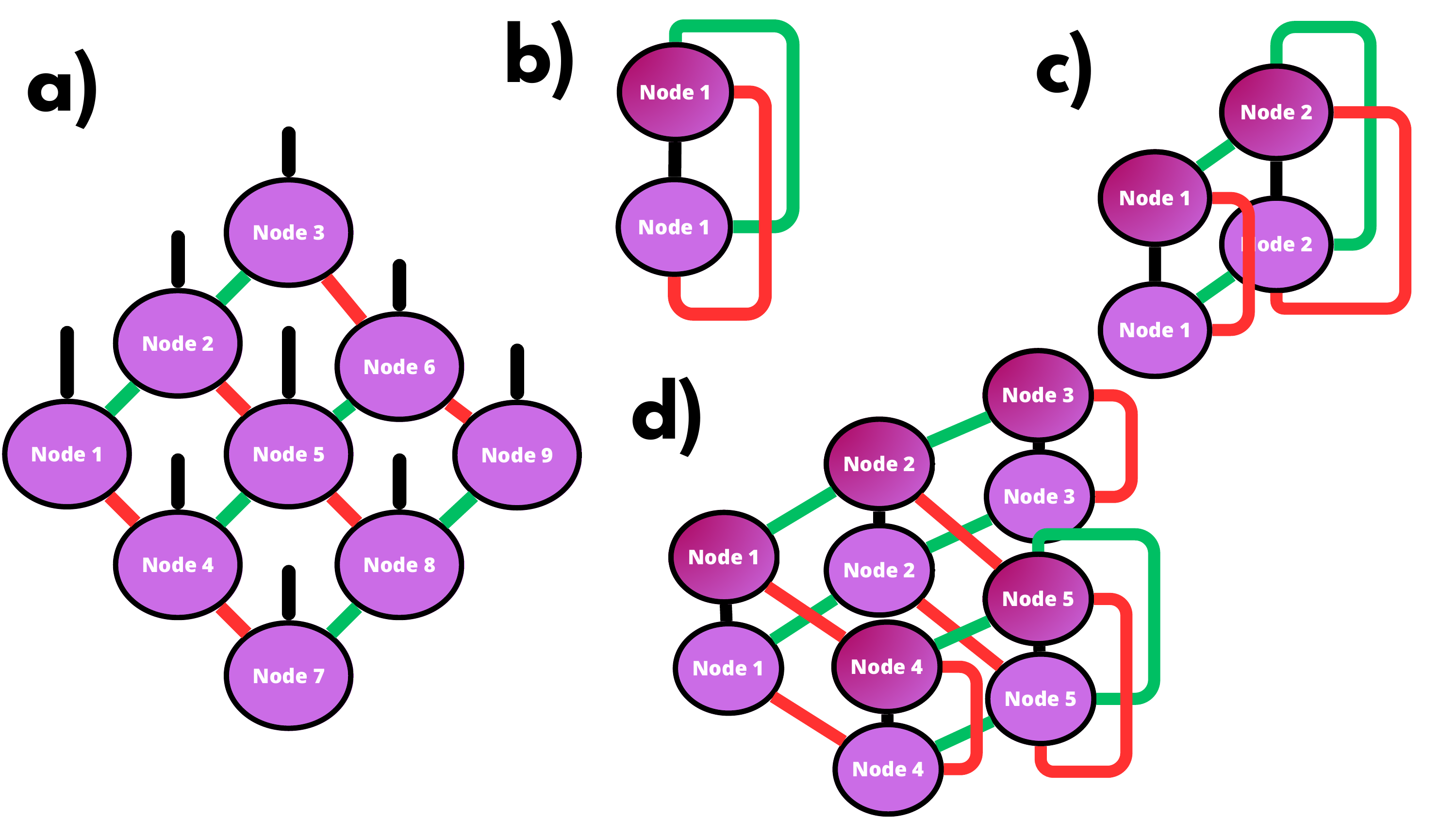}
    \caption{a) PEPS layer with 9 nodes. b) Partial square norm at 1 node. c) Partial square norm at 2 nodes. d) Partial square norm at 5 nodes.}
    \label{fig:PEPS}
\end{figure}

The same normalization principle applies to general tensor networks, but for networks with cycles, such as PEPS, the computational cost of the partial contractions depends on the contraction order, the induced boundary size/treewidth, and the bond dimensions. The efficient reuse discussed below is therefore directly guaranteed for the tensor network classes and contraction orders considered in this work.

\subsection{Frobenius Tensor Network Renormalization Initialization protocol}
If we have a tensor network $\mathcal{A}$, representing a $n_A\times m_A$ matrix whose Frobenius norm $||\mathcal{A}||_F$ is infinite (larger than our computer can store), zero (smaller than our computer precision), or outside a certain range of values, we will want to normalize the elements of our tensor network so that the norm $||\mathcal{B}||_F$ of the new tensor network $\mathcal{B}$ is equal to a given number $F>0$. We assume that the tensor represented by this tensor network has a smooth distribution in the values of its elements, due to a smooth distribution also of the elements of the nodes of its tensor network \cite{Gradient}. For a matrix of size $n_A\times m_A$ with entrywise scale of order one, a natural Frobenius target scales like $F=O(\sqrt{n_Am_A})$, although other heuristic targets can also be chosen.

If the total Frobenius norm $M=||\mathcal{A}||_F$ is finite and non-zero, exact normalization to the target $F$ is obtained by dividing each node by
\begin{equation}
    r = \left(\frac{M}{F}\right)^{1/N}.
\end{equation}
If instead we work with the closed contraction $S=||\mathcal{A}||_F^2$, the equivalent factor is
\begin{equation}
    r = \left(\frac{S}{F^2}\right)^{1/(2N)}.
\end{equation}
For the partial quantities, we write
\begin{equation}
    S_n = {}^{pF}||\mathcal{A}||_{n,N},
    \qquad
    S_n^* = F_n^2,
\end{equation}
where $F_n>0$ is a chosen target Frobenius norm for the $n$-core partial quantity. Exact normalization of a finite partial value uses
\begin{equation}
    r_n = \left(\frac{S_n}{S_n^*}\right)^{1/(2n)}.
\end{equation}
The tolerance window is therefore written as
\begin{equation}
    (\alpha F_n)^2 \leq S_n \leq (\beta F_n)^2,
\end{equation}
with $0<\alpha\leq 1\leq \beta$. If only the global target $F$ is prescribed, then the choice of partial targets $F_n$ should be understood explicitly as a heuristic auxiliary choice rather than as an exact consequence of the total target alone.

Since we cannot divide the elements by 0 or infinity, we use the following logic at the level of the computed floating-point sequence. If $x_1,\ldots,x_N$ denotes the computed sequence of partial square norms and $j$ is the first index such that $x_j$ is reported as $\infty$ or $0$, then, for $j>1$, $x_{j-1}$ was not reported with that same failure. This is a statement about the computed sequence, not about the exact tensor network; the case $j=1$ must be treated separately, and in tensor networks with signs the partial quantities need not be monotone.

The idea is to iteratively normalize the norm little by little so that we eventually achieve full normalization. To avoid having to start the iterative process from the beginning several times, we can take advantage of the fact that, if we normalize $\mathcal{A}_n$, all the subnetworks \mbox{$\mathcal{A}_m, \forall m<n$} that compose it will also be normalized. Because of this, each time we normalize, we will not have to start from the network with one node, but we can continue with the network with $n$ nodes.

For certain tensor networks, such as MPS, it is possible to reuse intermediate computations efficiently. For each normalization step, when we compute the partial norm, we first contract all the bond indexes that connect the nodes in the subnetwork and all the physical indexes. Before contracting the bond indexes connecting nodes that are not present, we store the tensor provisionally. Thus, after renormalization, we can normalize both the individual tensor elements and this provisional tensor so that we do not have to contract the entire tensor network again. Furthermore, we can use this tensor for the next steps of normalization.

We want a tensor network $\mathcal{A}$ with Frobenius norm $F$, with $N$ nodes, and we set partial targets $F_n$ together with tolerances $0<\alpha\leq 1\leq \beta$. The \textit{Frobenius Tensor Network Renormalization} initialization protocol to follow would be:
\begin{enumerate}
    \item We initialize the node tensors with some initialization method. We recommend random initialization with a Gaussian distribution of a constant standard deviation (not greater than $0.5$) and a constant mean neither too high nor too low and positive.
    \item We compute $M=||\mathcal{A}||_F$. If it is finite and non-zero, we divide each element of each node by $\left(\frac{M}{F}\right)^{1/N}$ and return $\mathcal{A}$. Otherwise, we continue.
    \item We compute ${}^{pF}||\mathcal{A}||_{1,N}$.
    \begin{enumerate}
        \item If it is infinite, we divide each element of the nodes of $\mathcal{A}$ by $(10(1+\xi))^{1/2}$, being $\xi$ a random number between 0 and 1, and return to Step 2.
        \item If it is zero, we multiply each element of the nodes of $\mathcal{A}$ by $(10(1+\xi))^{1/2}$ and return to Step 2.
        \item Otherwise, we save this value as ${}^{pF}||\mathcal{A}||_{1,N}$ and continue.
      \end{enumerate}
  
    \item For $n\in [2, N-1]$, we compute ${}^{pF}||\mathcal{A}||_{n,N}$.
    \begin{enumerate}
    \item If it is infinite, we divide each element of the nodes of $\mathcal{A}$ by $\max\left\{\left(\frac{{}^{pF}||\mathcal{A}||_{n-1,N}}{F_{n-1}^2}\right)^{\frac{1}{2(n-1)}}, (10(1+\xi))^{\frac{1}{2(n-1)}}\right\}$, and repeat Steps 2 and 4 (from this value of $n$).
    \item If it is zero, we multiply each element of the nodes of $\mathcal{A}$ by $\max\left\{\left(\frac{F_{n-1}^2}{{}^{pF}||\mathcal{A}||_{n-1,N}}\right)^{\frac{1}{2(n-1)}}, (10(1+\xi))^{\frac{1}{2(n-1)}}\right\}$, and repeat Steps 2 and 4 (from this value of $n$).
    \item If it is finite, but outside the interval $\left[(\alpha F_n)^2,(\beta F_n)^2\right]$, we divide each element of the nodes of $\mathcal{A}$ by $\left(\frac{{}^{pF}||\mathcal{A}||_{n,N}}{F_n^2}\right)^{\frac{1}{2n}}$, and repeat Steps 2 and 4 (from this value of $n$).
    \item Otherwise, we continue.
    \end{enumerate}
    
    \item If no partial square norm is outside the range, infinite or zero, but the total norm is still unavailable or outside tolerance, we divide each element of the nodes of $\mathcal{A}$ by $\left(\frac{{}^{pF}||\mathcal{A}||_{N-1,N}}{F_{N-1}^2}\right)^{\frac{1}{2(N-1)}}$, and repeat Steps 2 and 5. This last step is a fallback based on the last available partial quantity.
\end{enumerate}

We repeat the cycle until we obtain a valid $\mathcal{A}$ or we reach a stop condition, which entails repeating a certain maximum number of iterations. If we reach this last point, the protocol will have failed and we will have two options. The first is to change the order of the nodes so that other structures are checked. The second is to reinitialize with other hyperparameters in the initialization protocol.

The purpose of using a random factor in the case of divergence in the partial norm with one node is that, not knowing the real value by which we should divide or multiply, we rescale by an order of magnitude in the correct direction. However, to avoid possible infinite rescaling loops, we add a variability factor so that we cannot get stuck.

In case the elements of certain tensors of the tensor network have values smaller than or larger than those of the other tensors, there is a possibility that instead of evenly distributing renormalization among the $n$ cores involved in a partial correction, we allow the other nodes to renormalize differently, equalizing the proportion of the values. We can also try to have as many integer values as possible in the values of the tensor elements by an adjusted renormalization of each tensor, or perform a quantization. These two processes can be performed either before, during, or after finishing the renormalization process, it being advisable to perform it at least at the end.

\section{Positive Tensor network initialization protocol LTNR}\label{sec: positive initial}
This case deals with tensor networks whose represented entries are nonnegative, which allows us to apply computationally less expensive techniques. Our protocol is based on the use of a partial positive lineal entrywise sum to normalize the total positive lineal entrywise sum of the resulting tensor.

\begin{definition}[Positive lineal entrywise sum]
$ $
\\
Let $A$ be a matrix with $a_{ij}\geq 0$ for all $i,j$. The positive lineal entrywise sum of $A$ is given by 
\begin{equation}
    ||A||_L = \sum_{ij}a_{ij} =  \mathbbm{1}^T A\mathbbm{1},
    \label{eq: Lineal}
\end{equation}
being $\mathbbm{1}=(1,1,\dots)$ a ones vector. We keep the notation $||A||_L$ and the acronym LTNR for continuity, but this quantity is not a norm on arbitrary real matrices because cancellations may make it vanish and negative scalars break homogeneity.
\end{definition}

\begin{figure}[h]
    \centering
    \includegraphics[width=0.45\textwidth]{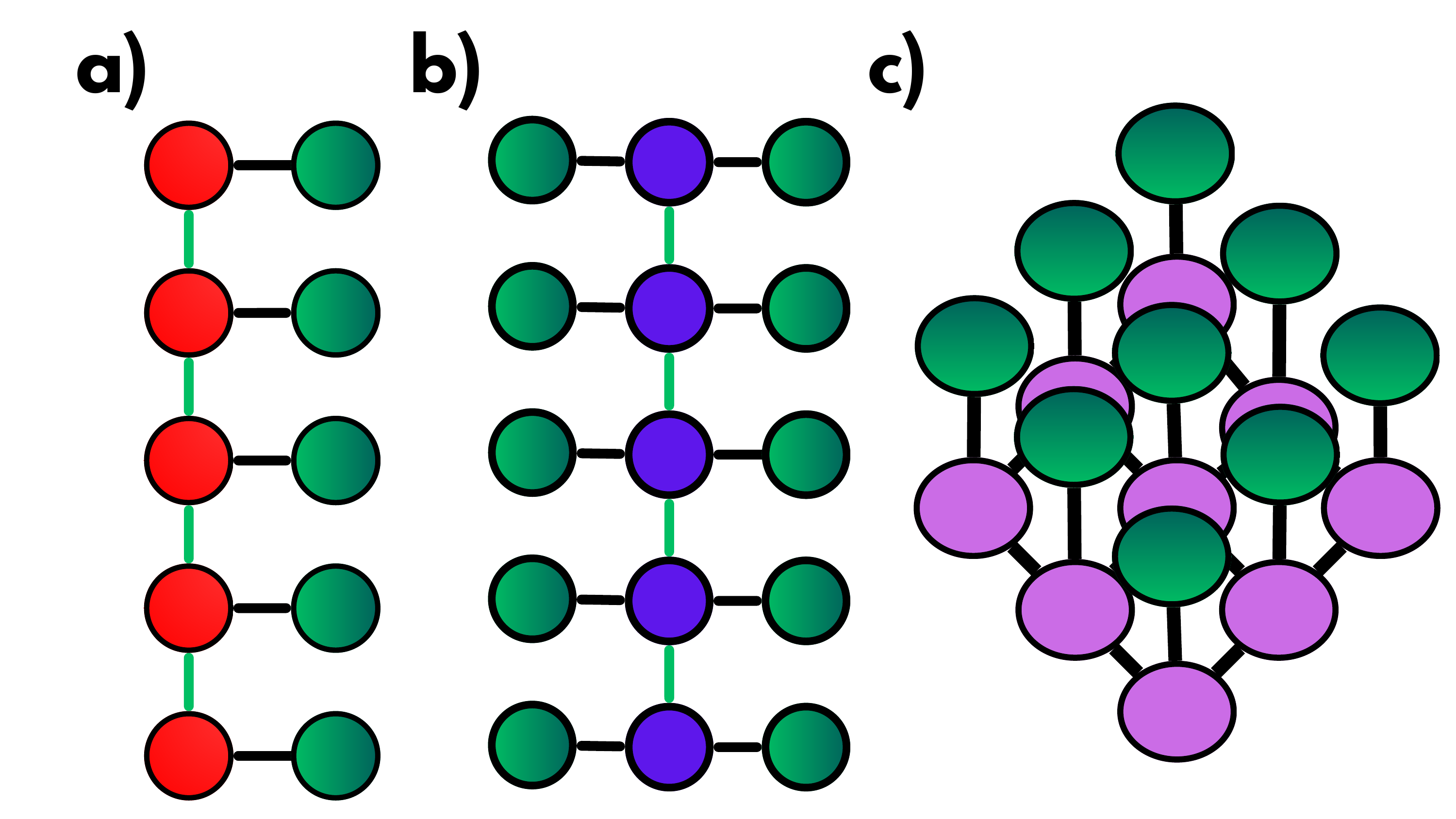}
    \caption{Positive lineal entrywise sum calculated to a) Tensor Train layer (red nodes). b) Tensor Train Matrix layer (blue nodes). c) PEPS layer (purple nodes). The green nodes are ones vectors.}
    \label{fig:Lineal_tot}
\end{figure}

In a tensor network, this would be to contract the layer with a set of nodes of ones vectors. We can see some examples in Fig. \ref{fig:Lineal_tot}. The contraction of this tensor network is equivalent to summing all the elements of the matrix it represents. As in the case of the Frobenius norm, it can be computed without the need to calculate the elements of the represented matrix, using only the elements of the nodes.

In this case, we can interpret this rule in a manner analogous to the Frobenius norm in the general case, since all the elements are nonnegative, we will have that
\begin{equation}
    ||A||_L = \sum_{ij}a_{ij} = \sum_{ij}|\sqrt{a_{ij}}|^2 = ||A^{\circ \frac{1}{2}}||_F^2,
\end{equation}
assuming $A\geq 0$ entrywise. We also have
\begin{equation}
    ||A^{\circ \frac{1}{2}}||_F = \sqrt{\sum_{ij}a_{ij}}.
\end{equation}
Therefore, for nonnegative matrices, the positive entrywise sum is equal to the square of the Frobenius norm of the Hadamard root of the matrix, not to the Frobenius norm itself. With a smooth distribution of elements around $a_{00}$, the quantity in Eq. \eqref{eq: Lineal} will be of the order of $nma_{00}$ for a $n\times m$ matrix.

As in the general case, the way to avoid values that are too large or too small is to normalize the matrix, but this time with respect to the positive lineal entrywise sum. For a matrix $n \times m$, this quantity naturally scales like $nm$ when the mean entrywise value stays of order one. Moreover, if all tensor cores have nonnegative entries and the contraction uses only products and sums, then the represented tensor or matrix is entrywise nonnegative.

For this purpose, we define what we will call the partial positive lineal sum of the tensor network, analogous to the partial square norm of the Frobenius norm method.

\subsection{Partial positive lineal sum of the tensor network}
\begin{definition}[Partial positive lineal sum]
$ $
\\
Given a tensor network $\mathcal{A}$ with $N$ nodes, and the tensor network $\mathcal{A}_n$ defined by the first $n$ nodes of $\mathcal{A}$, we define ${}^{pL}||\mathcal{A}||_{n,N}$, the partial positive lineal sum at $n$ nodes of $\mathcal{A}$ as the positive lineal entrywise sum of $\mathcal{A}_n$. That is,
\begin{equation}
    {}^{pL}||\mathcal{A}||_{n,N} = ||\mathcal{A}_n||_L.
\end{equation}
\end{definition}

\begin{figure}[h]
    \centering
    \includegraphics[width=0.45\textwidth]{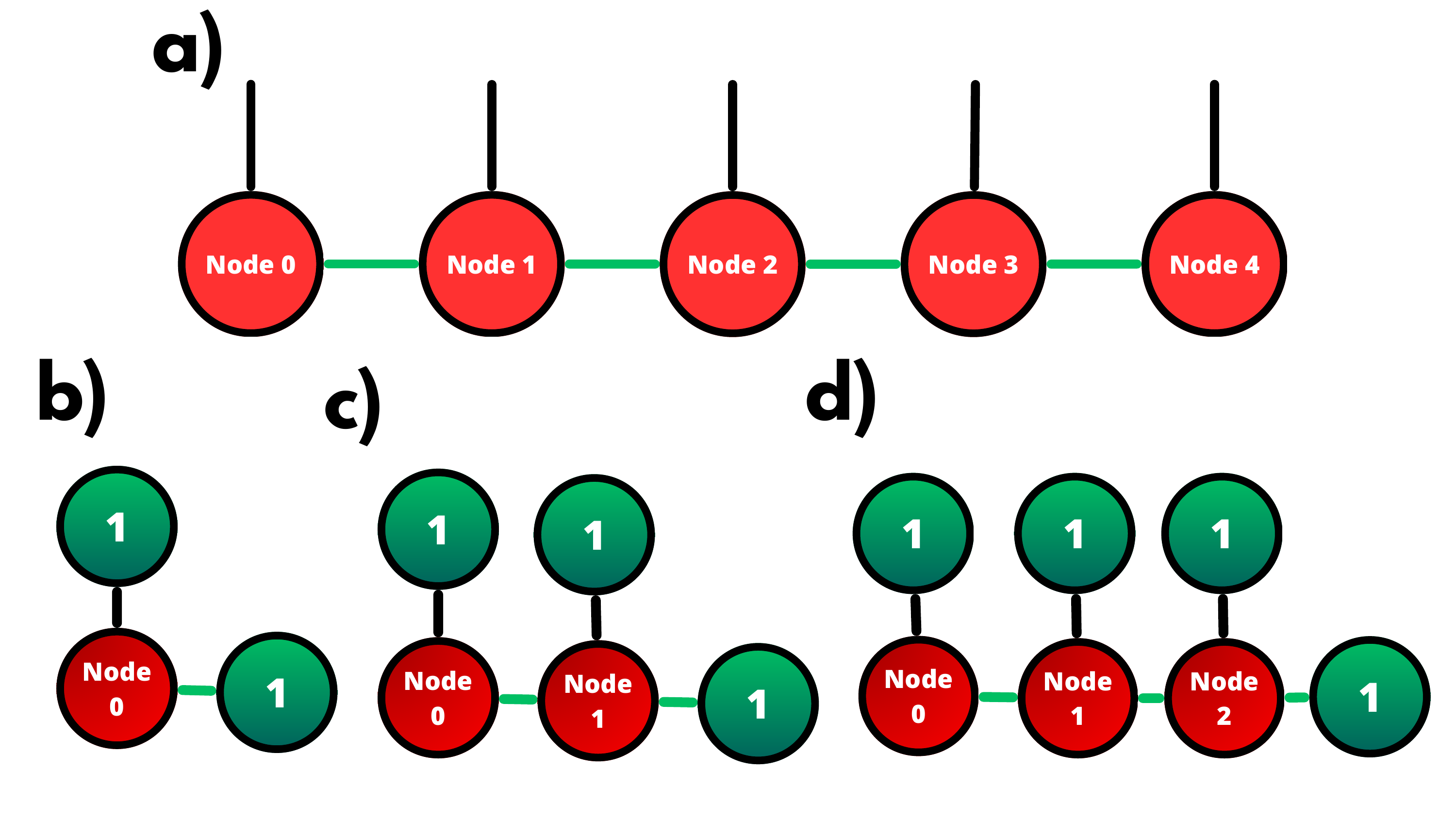}
    \caption{a) Tensor Train layer with 5 nodes. b) Partial positive lineal sum at 1 node. c) Partial positive lineal sum at 2 nodes. d) Partial positive lineal sum at 3 nodes. The 1 nodes are ones vectors.}
    \label{fig:TT_Partial_Pos}
\end{figure}

To get an idea of what this partial positive lineal sum is, we will exemplify it with a simple case, a tensor train layer. We will consider the tensor network in Fig. \ref{fig:TT_Partial_Pos}, whose nodes are sorted.

As we can see, in this case we would only have to do the same process as when calculating the total norm of the total tensor network, but stop at step $n$ and contract the bond index of the final tensor of the chain with a ones vector.

We can see in the following Fig. \ref{fig:TT_Matrix_Pos} and \ref{fig:PEPS_Pos} how the partial positive lineal sum would be for a TT-Matrix layer and for a PEPS layer. The same normalization principle extends to general tensor networks, although for networks with cycles the computational cost again depends on the contraction order and the induced boundary size.
\begin{figure}[h]
    \centering
    \includegraphics[width=0.45\textwidth]{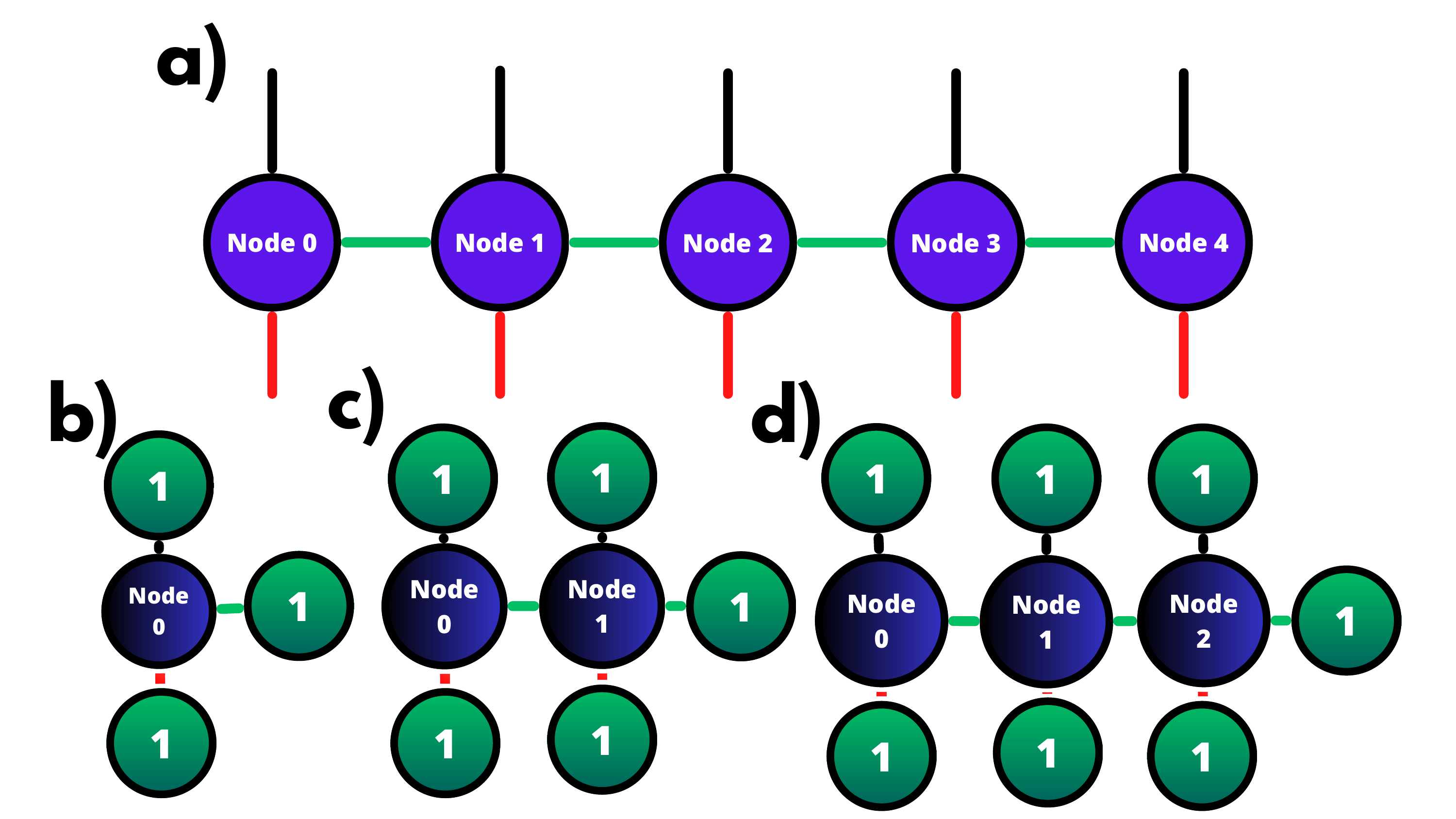}
    \caption{a) Tensor Train Matrix layer with 5 nodes. b) Partial positive lineal sum at 1 node. c) Partial positive lineal sum at 2 nodes. d) Partial positive lineal sum at 3 nodes. The 1 nodes are ones vectors.}
    \label{fig:TT_Matrix_Pos}
\end{figure}
\begin{figure}[h]
    \centering
    \includegraphics[width=0.45\textwidth]{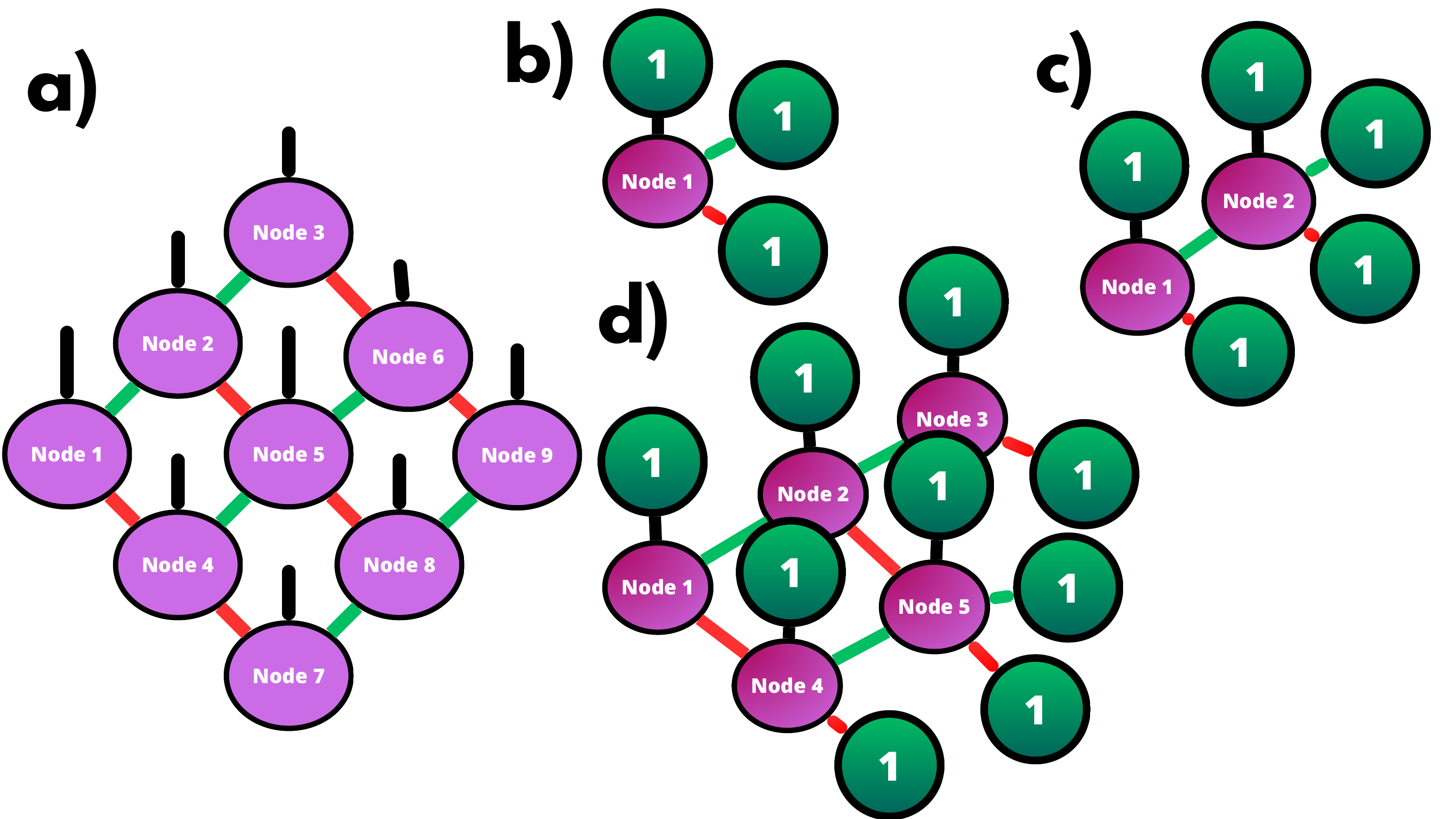}
    \caption{a) PEPS layer with 9 nodes. b) Partial positive lineal sum at 1 node. c) Partial positive lineal sum at 2 nodes. d) Partial positive lineal sum at 5 nodes. The 1 nodes are ones vectors.}
    \label{fig:PEPS_Pos}
\end{figure}

\subsection{Lineal Tensor Network Renormalization initialization protocol}
If we have a tensor network $\mathcal{A}$, representing a nonnegative matrix $n_A\times m_A$ whose positive lineal entrywise sum $||\mathcal{A}||_L$ is infinite, zero or outside a certain range of values, we will want to normalize the elements of our tensor network so that the quantity $||\mathcal{B}||_L$ of the new tensor network $\mathcal{B}$ is equal to a certain number $L^*>0$. We assume that the tensor represented by this tensor network has a smooth distribution in the values of its elements, due to a smooth distribution also of the elements of the nodes of its tensor network \cite{Gradient}. For a matrix of size $n_A\times m_A$ with entrywise scale of order one, a natural target for this quantity scales like $L^*=O(n_Am_A)$.

If the total positive lineal entrywise sum $L=||\mathcal{A}||_L$ is finite and non-zero, exact normalization to the target $L^*$ is obtained by dividing each node by
\begin{equation}
    r = \left(\frac{L}{L^*}\right)^{1/N}.
\end{equation}
For the partial quantities, we write
\begin{equation}
    L_n = {}^{pL}||\mathcal{A}||_{n,N},
    \qquad
    L_n^*>0,
\end{equation}
and exact normalization of a finite partial value uses
\begin{equation}
    r_n = \left(\frac{L_n}{L_n^*}\right)^{1/n}.
\end{equation}
The partial tolerance window is therefore
\begin{equation}
    \alpha L_n^* \leq L_n \leq \beta L_n^*,
\end{equation}
with $0<\alpha\leq 1\leq \beta$.

The same reasons as in the previous algorithm led to an analogous algorithm, but with the change from the partial square norm to the partial positive lineal sum. The idea is to iteratively normalize the quantity little by little so that we eventually achieve full normalization. As in the general case, after each normalization, there is no need to recalculate from a single node, and you can continue from the current subnetwork.

We want a tensor network $\mathcal{A}$ with positive lineal entrywise sum $L^*$, with $N$ nodes, and we set partial targets $L_n^*$ together with tolerances $0<\alpha\leq 1\leq \beta$. The \textit{Lineal Tensor Network Renormalization} initialization protocol to follow would be:
\begin{enumerate}
    \item We initialize the node tensors with some initialization method that preserves nonnegativity, for example a positive uniform distribution, a lognormal distribution, a gamma distribution, a truncated normal on positive values, or the absolute value of a Gaussian.
    \item We compute $L=||\mathcal{A}||_L$. If it is finite and non-zero, we divide each element of each node by $\left(\frac{L}{L^*}\right)^{1/N}$ and return $\mathcal{A}$. Otherwise, we continue.
    \item We compute ${}^{pL}||\mathcal{A}||_{1,N}$.
    \begin{enumerate}
        \item If it is infinite, we divide each element of the nodes of $\mathcal{A}$ by $10(1+\xi)$, being $\xi$ a random number between 0 and 1, and return to Step 2.
        \item If it is zero, we multiply each element of the nodes of $\mathcal{A}$ by $10(1+\xi)$ and return to Step 2.
        \item Otherwise, we save this value as ${}^{pL}||\mathcal{A}||_{1,N}$ and continue.
      \end{enumerate}
  
    \item For $n\in [2, N-1]$, we compute ${}^{pL}||\mathcal{A}||_{n,N}$.
    \begin{enumerate}
    \item If it is infinite, we divide each element of the nodes of $\mathcal{A}$ by $\max\left\{\left(\frac{{}^{pL}||\mathcal{A}||_{n-1,N}}{L_{n-1}^*}\right)^{\frac{1}{n-1}}, (10(1+\xi))^{\frac{1}{n-1}}\right\}$, and repeat Steps 2 and 4 (from this value of $n$).
    \item If it is zero, we multiply each element of the nodes of $\mathcal{A}$ by $\max\left\{\left(\frac{L_{n-1}^*}{{}^{pL}||\mathcal{A}||_{n-1,N}}\right)^{\frac{1}{n-1}}, (10(1+\xi))^{\frac{1}{n-1}}\right\}$, and repeat Steps 2 and 4 (from this value of $n$).
    \item If it is finite, but outside the interval $\left[\alpha L_n^*,\beta L_n^*\right]$, we divide each element of the nodes of $\mathcal{A}$ by $\left(\frac{{}^{pL}||\mathcal{A}||_{n,N}}{L_n^*}\right)^{\frac{1}{n}}$, and repeat Steps 2 and 4 (from this value of $n$).
    \item Otherwise, we continue.
    \end{enumerate}
    
    \item If no partial positive lineal sum is outside the range, infinite or zero, but the total quantity is still unavailable or outside tolerance, we divide each element of the nodes of $\mathcal{A}$ by $\left(\frac{{}^{pL}||\mathcal{A}||_{N-1,N}}{L_{N-1}^*}\right)^{\frac{1}{N-1}}$, and repeat Steps 2 and 5. This last step is a fallback based on the last available partial quantity. 
\end{enumerate}

As in the general case, we repeat the cycle until we reach a stop condition, which will be to have repeated a certain maximum number of iterations. If we reach that point, the protocol will have failed, and we will have the same two options as before.

\section{Experiments}\label{sec: experiments}
In this section, we will perform several experiments with both algorithms. We will check the scaling of the number of steps needed to normalize the tensor network, considering one step each time we have to partially normalize because we have found an infinity or zero in the partial norms. We tested for the TT layer and the TT matrix layer with $N$ nodes, uniform physical dimensions $p$ and bond dimensions $b$. For FTNR, we use a Gaussian initialization with mean $1$ and standard deviation $0.5$. For LTNR, we use the absolute value of that Gaussian initialization so that the entries are nonnegative. We choose $F=p^{N}$. For the TT-matrix layer, this has the same scaling as the square root of the number of entries, while for the TT layer it coincides with the number of entries and should therefore be interpreted as a heuristic target rather than the canonical Frobenius scaling $O(\sqrt{p^N})$. Our tolerance range for the total target is $(10^{-3}F ,10^{3}F)$.

\begin{figure}[h]
    \centering
    \includegraphics[width=\linewidth]{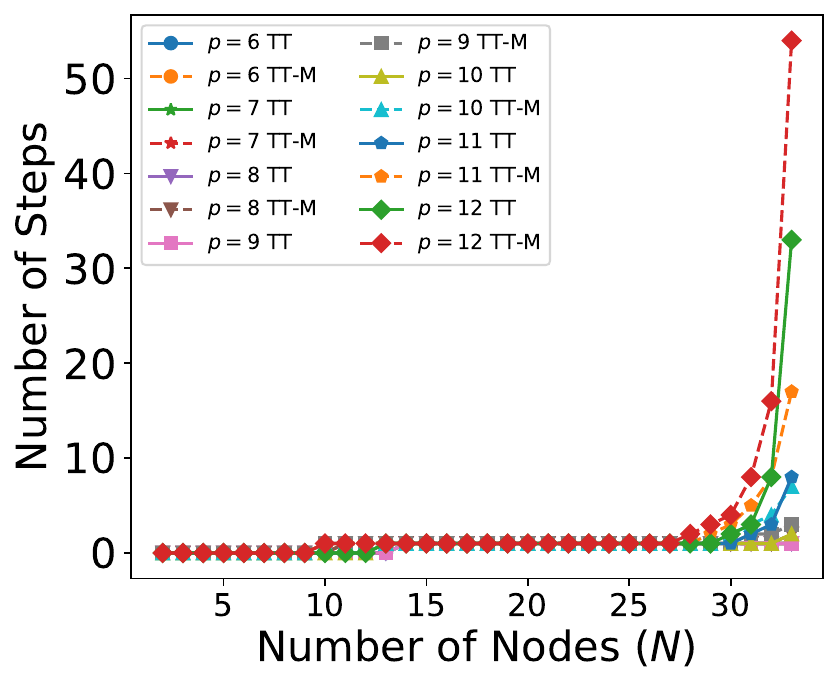}
    \caption{Number of steps vs $N$ for $p$ from 6 to 12 for the TT layer and TT-Matrix layer with fixed $b=12$ for the Frobenius method.}
    \label{fig:StepsFrob}
\end{figure}

\begin{figure}[h]
    \centering
    \includegraphics[width=\linewidth]{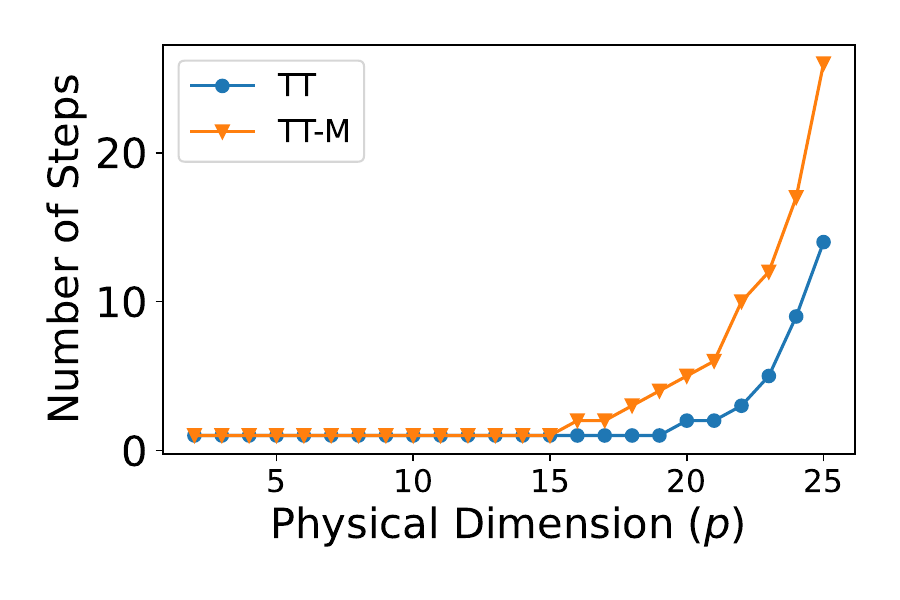}
    \caption{Number of steps vs $p$ with fixed $N=25$ and $b=10$ for the TT layer and the TT-Matrix layer for the Frobenius method.}
    \label{fig:Steps2Frob}
\end{figure}

\begin{figure}[h]
    \centering
    \includegraphics[width=\linewidth]{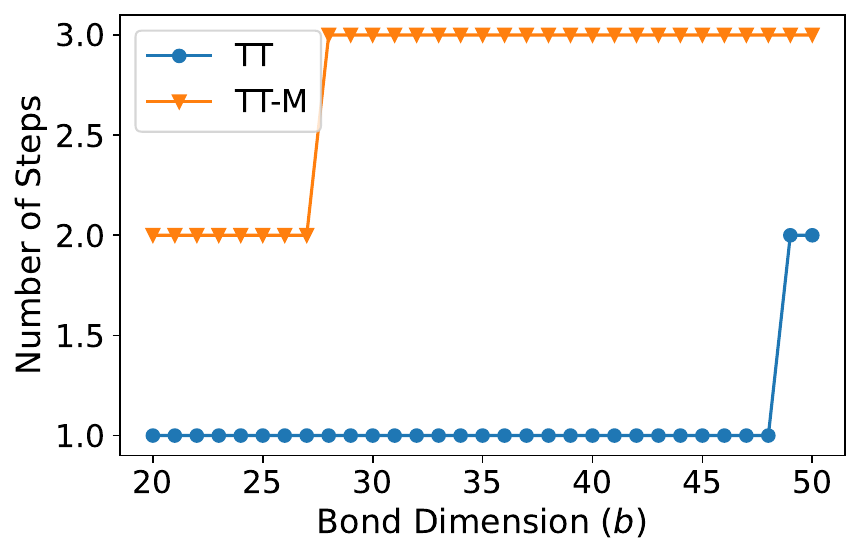}
    \caption{Number of steps vs $b$ with fixed $N=25$ and $p=15$ for the TT layer and the TT-Matrix layer for the Frobenius method.}
    \label{fig:Steps3Frob}
\end{figure}

We first test the performance of the normalization with the Frobenius norm. First, we check the number of steps versus the number $N$ of the nodes of the tensor network, from $2$ to $34$, for different values of $p$ with $b=12$ in Fig. \ref{fig:StepsFrob}. Then, we check the number of steps against $p$ for the same value of $N=25$ and $b=10$ in Fig. \ref{fig:Steps2Frob}. Finally, we check the number of steps against $b$ with the same value of $N=25$ and $p=15$ in Fig. \ref{fig:Steps3Frob}.


We can observe that for all cases, for $N<10$ no steps are needed, while there is a region up to $N=27$ where only one step is needed. For a larger number of nodes, the number of steps increases exponentially with $N$, increasing faster for larger $p$. Against $p$, we see that for $p<15$ only one step is needed, while for larger values, there is another exponential growth. Relative to $b$, no substantial dependence is observed. The algorithm should need more steps for larger cases, but we have not enough memory to compute them. For all checks, the TT-Matrix requires a larger number of steps.

\begin{figure}[h]
    \centering
    \includegraphics[width=\linewidth]{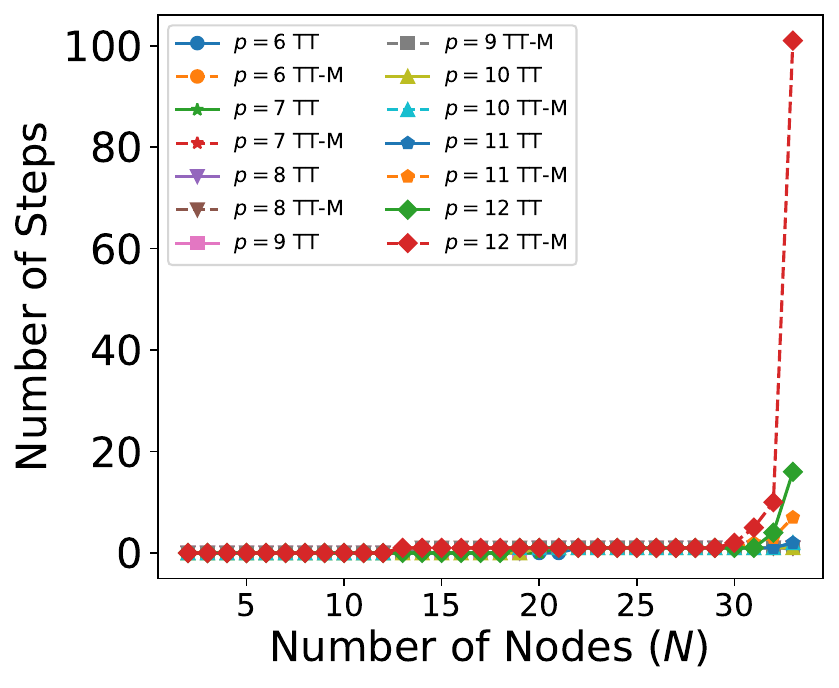}
    \caption{Number of steps vs $N$ for $p$ from 6 to 12 for the TT layer and TT-Matrix layer with fixed $b=10$ for the lineal method.}
    \label{fig:StepsLineal}
\end{figure}

\begin{figure}[h]
    \centering
    \includegraphics[width=\linewidth]{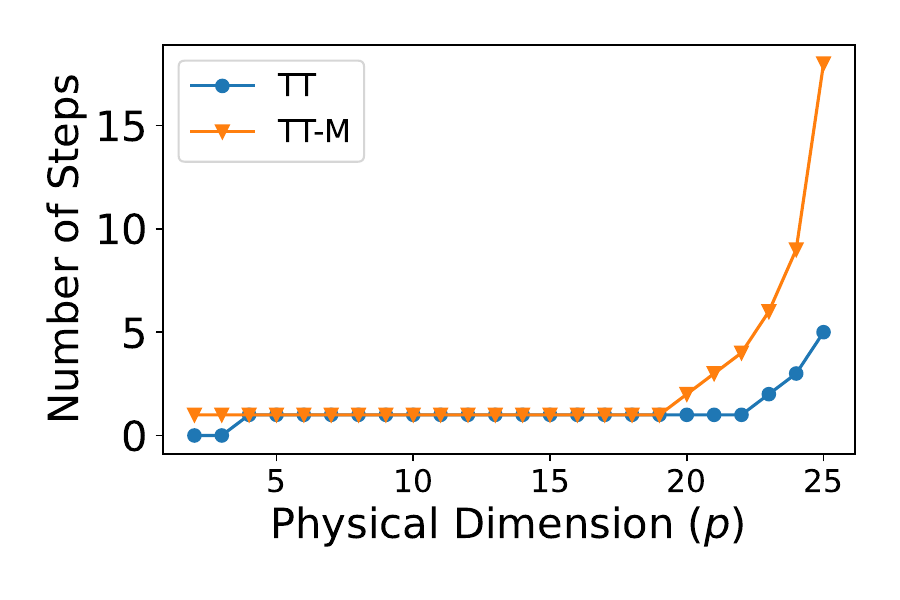}
    \caption{Number of steps vs $p$ with fixed $N=25$ and $b=10$ for the TT layer and the TT-Matrix layer for the lineal method.}
    \label{fig:Steps2Lineal}
\end{figure}

\begin{figure}[h]
    \centering
    \includegraphics[width=\linewidth]{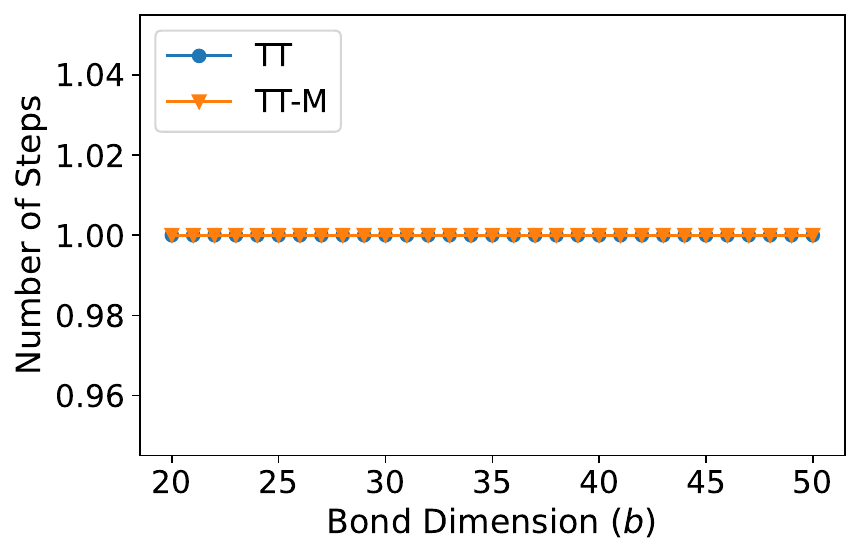}
    \caption{Number of steps vs $b$ with fixed $N=25$ and $p=15$ for the TT layer and the TT-Matrix layer for the lineal method.}
    \label{fig:Steps3Lineal}
\end{figure}

Now, we test the LTNR algorithm, with the same configurations as in the Frobenius experiment. In Fig.~\ref{fig:StepsLineal}, we can observe a similar tendence as in the Frobenius case. All instances of $N<13$ need no steps and for $N<30$ only one step is enough to normalize the matrix. In the following region, an exponential amount of steps are needed, but not as much as in the Frobenius case. Only the $p=12$ TT-matrix instance needs more steps. In Fig.~\ref{fig:Steps2Lineal}, we can observe the same behavior as in the Frobenius case, but with less required steps. Finally, in Fig.~\ref{fig:Steps3Lineal}, there is no clear dependence for the number of steps and $b$, probably for the same reasons as before.


As we can see, both algorithms perform notably well in a wide range of possible instance sizes. Moreover, the lineal algorithm performs better, probably due to the smoother scaling of the positive lineal entrywise sum of the tensor, which grows linearly with the number of entries, while the Frobenius norm is the square root of a quadratic contraction.

\section{Other applications}
\label{sec:other}
So far we have seen the application of tensorized neural networks, but this method can be useful for more fields. Every algorithm or application that needs to contract a tensor network and the non-zero elements of the tensors that form it are of the same order of magnitude, we can benefit from these methods. This can be helpful in cases where we do not want the absolute scale of the final tensor elements but we want to observe a relative scale between them.

An example would be the simulation of imaginary time evolution processes where we want to see which is the state with minimum energy and not the energy it has. However, this energy could be recovered if we performed the method, saved the scale factor by which we multiply the elements of the tensor network, and multiplied the values of the resulting tensor network by this factor. This can be interesting because the different factors can be multiplied keeping their order of magnitude apart, so that we do not have overflows.

\section{Conclusions}\label{sec: conclusions}
We have developed two methods to successfully initialize layers of tensorized neural networks using partial computations of their Frobenius norms and positive lineal entrywise sums, depending on the type of tensor network involved. We have exposed a way to take advantage of intermediate calculations for various types of tensor network to optimize this process. We have also applied it to different layers and seen its scaling versus the number of nodes, bond dimension, and physical dimension.

The main limitation of these methods is the need for the values of the represented tensor elements to follow a smooth distribution, which can be limiting in certain cases. This limitation must be overcomed in order to generalize the scope of application of this kind of algorithms, allowing new tensor network algorithms to be possible. In addition, the present experiments mainly assess the initial normalization/stabilization stage; by themselves they do not establish training-gradient stability, predictive-performance gains, seed robustness, full entrywise concentration, or time/memory optimality against alternative initializations.

A possible future line of research could be to investigate how to reduce the number of steps to be performed. Another could be to study the scaling of complexity with increasing size of each of the different types of existing layers. We could also apply it to the methods mentioned in Sec. \ref{sec:other}, for example in combinatorial optimization \cite{Optimiz} to determine the appropriate decay factor and adapt it to quantum machine learning layers.

\nocite{*}

\bibliography{apssamp}% Produces the bibliography via BibTeX.

@PREAMBLE{
 "\providecommand{\noopsort}[1]{}" 
 # "\providecommand{\singleletter}[1]{#1}%" 
}

@misc{Tensor,
      title={Tensor Networks in a Nutshell}, 
      author={Jacob Biamonte and Ville Bergholm},
      year={2017},
      eprint={1708.00006},
      archivePrefix={arXiv}
}

@misc{Tensorizing,
      title={Tensorizing Neural Networks}, 
      author={Alexander Novikov and Dmitry Podoprikhin and Anton Osokin and Dmitry Vetrov},
      year={2015},
      eprint={1509.06569},
      archivePrefix={arXiv}
}

@misc{LowRank,
      title={Exploiting Low-Rank Tensor-Train Deep Neural Networks Based on Riemannian Gradient Descent With Illustrations of Speech Processing}, 
      author={Jun Qi and Chao-Han Huck Yang and Pin-Yu Chen and Javier Tejedor},
      year={2022},
      eprint={2203.06031},
      archivePrefix={arXiv}
}

@misc{Convolutional,
      title={Deep convolutional tensor network}, 
      author={Philip Blagoveschensky and Anh Huy Phan},
      year={2020},
      eprint={2005.14506},
      archivePrefix={arXiv}
}

@misc{Anomaly,
      title={Anomaly Detection with Tensor Networks}, 
      author={Jinhui Wang and Chase Roberts and Guifre Vidal and Stefan Leichenauer},
      year={2020},
      eprint={2006.02516},
      archivePrefix={arXiv}
}

@article{Optimiz,
   title={A Quantum-Inspired Tensor Network Algorithm for Constrained Combinatorial Optimization Problems},
   volume={10},
   ISSN={2296-424X},
   url={http://dx.doi.org/10.3389/fphy.2022.906590},
   DOI={10.3389/fphy.2022.906590},
   journal={Frontiers in Physics},
   publisher={Frontiers Media SA},
   author={Hao, Tianyi and Huang, Xuxin and Jia, Chunjing and Peng, Cheng},
   year={2022},
   month=jul }

@misc{DeepLearning,
      title={A Survey on Deep Learning and State-of-the-art Applications}, 
      author={Mohd Halim Mohd Noor and Ayokunle Olalekan Ige},
      year={2024},
      eprint={2403.17561},
      archivePrefix={arXiv}
}

@misc{TNNPhysics,
      title={Quantum-Inspired Tensor Neural Networks for Partial Differential Equations}, 
      author={Raj Patel and Chia-Wei Hsing and Serkan Sahin and Saeed S. Jahromi and Samuel Palmer and Shivam Sharma and Christophe Michel and Vincent Porte and Mustafa Abid and Stephane Aubert and Pierre Castellani and Chi-Guhn Lee and Samuel Mugel and Roman Orus},
      year={2022},
      eprint={2208.02235},
      archivePrefix={arXiv}
}

@misc{Gradient,
      title={Improvements to Gradient Descent Methods for Quantum Tensor Network Machine Learning}, 
      author={Fergus Barratt and James Dborin and Lewis Wright},
      year={2022},
      eprint={2203.03366},
      archivePrefix={arXiv}
}

@misc{MPS1,
      title={Matrix Product State Representations}, 
      author={D. Perez-Garcia and F. Verstraete and M. M. Wolf and J. I. Cirac},
      year={2007},
      eprint={quant-ph/0608197},
      archivePrefix={arXiv},
      primaryClass={quant-ph}
}

@article{MPS2,
author = {F. Verstraete, V. Murg and J.I. Cirac},
title = {Matrix product states, projected entangled pair states, and variational renormalization group methods for quantum spin systems},
journal = {Advances in Physics},
volume = {57},
number = {2},
pages = {143-224},
year = {2008},
publisher = {Taylor & Francis},
doi = {10.1080/14789940801912366},

URL = { 
        https://doi.org/10.1080/14789940801912366
},
eprint = { 
    
        https://doi.org/10.1080/14789940801912366

}

}

@misc{LLaMa,
      title={LLaMA: Open and Efficient Foundation Language Models}, 
      author={Hugo Touvron and Thibaut Lavril and Gautier Izacard and Xavier Martinet and Marie-Anne Lachaux and Timothée Lacroix and Baptiste Rozière and Naman Goyal and Eric Hambro and Faisal Azhar and Aurelien Rodriguez and Armand Joulin and Edouard Grave and Guillaume Lample},
      year={2023},
      eprint={2302.13971},
      archivePrefix={arXiv},
      primaryClass={cs.CL},
      url={https://arxiv.org/abs/2302.13971}, 
}

@article{Quantum_Compression,
   title={Guided quantum compression for high dimensional data classification},
   volume={5},
   ISSN={2632-2153},
   url={http://dx.doi.org/10.1088/2632-2153/ad5fdd},
   DOI={10.1088/2632-2153/ad5fdd},
   number={3},
   journal={Machine Learning: Science and Technology},
   publisher={IOP Publishing},
   author={Belis, Vasilis and Odagiu, Patrick and Grossi, Michele and Reiter, Florentin and Dissertori, Günther and Vallecorsa, Sofia},
   year={2024},
   month=jul, pages={035010} }

@article{OrusTN,
   title={A practical introduction to tensor networks: Matrix product states and projected entangled pair states},
   volume={349},
   ISSN={0003-4916},
   url={http://dx.doi.org/10.1016/j.aop.2014.06.013},
   DOI={10.1016/j.aop.2014.06.013},
   journal={Annals of Physics},
   publisher={Elsevier BV},
   author={Orús, Román},
   year={2014},
   month=oct, pages={117–158} }

@article{NN_Compression,
   title={Compressing Neural Networks Using Tensor Networks with Exponentially Fewer Variational Parameters},
   volume={4},
   ISSN={2771-5892},
   url={http://dx.doi.org/10.34133/icomputing.0123},
   DOI={10.34133/icomputing.0123},
   journal={Intelligent Computing},
   publisher={American Association for the Advancement of Science (AAAS)},
   author={Qing, Yong and Li, Ke and Zhou, Peng-Fei and Ran, Shi-Ju},
   year={2025},
   month=jan }

@misc{convo_compression,
      title={Tensor network compressibility of convolutional models}, 
      author={Sukhbinder Singh and Saeed S. Jahromi and Roman Orus},
      year={2024},
      eprint={2403.14379},
      archivePrefix={arXiv},
      primaryClass={cs.CV},
      url={https://arxiv.org/abs/2403.14379}, 
}

@article{Transformer_compress,
title = {T3SRS: Tensor Train Transformer for compressing sequential recommender systems},
journal = {Expert Systems with Applications},
volume = {238},
pages = {122260},
year = {2024},
issn = {0957-4174},
doi = {https://doi.org/10.1016/j.eswa.2023.122260},
url = {https://www.sciencedirect.com/science/article/pii/S0957417423027628},
author = {Hao Li and Jianli Zhao and Huan Huo and Sheng Fang and Jianjian Chen and Lutong Yao and Yiran Hua}
}

@misc{LLM_quantum_tn,
      title={Quantum Large Language Models via Tensor Network Disentanglers}, 
      author={Borja Aizpurua and Saeed S. Jahromi and Sukhbinder Singh and Roman Orus},
      year={2024},
      eprint={2410.17397},
      archivePrefix={arXiv},
      primaryClass={quant-ph},
      url={https://arxiv.org/abs/2410.17397}, 
}

@misc{Compactifai,
      title={CompactifAI: Extreme Compression of Large Language Models using Quantum-Inspired Tensor Networks}, 
      author={Andrei Tomut and Saeed S. Jahromi and Abhijoy Sarkar and Uygar Kurt and Sukhbinder Singh and Faysal Ishtiaq and Cesar Muñoz and Prabdeep Singh Bajaj and Ali Elborady and Gianni del Bimbo and Mehrazin Alizadeh and David Montero and Pablo Martin-Ramiro and Muhammad Ibrahim and Oussama Tahiri Alaoui and John Malcolm and Samuel Mugel and Roman Orus},
      year={2024},
      eprint={2401.14109},
      archivePrefix={arXiv},
      primaryClass={cs.CL},
      url={https://arxiv.org/abs/2401.14109}, 
}

@misc{snn_tn,
      title={TT-SNN: Tensor Train Decomposition for Efficient Spiking Neural Network Training}, 
      author={Donghyun Lee and Ruokai Yin and Youngeun Kim and Abhishek Moitra and Yuhang Li and Priyadarshini Panda},
      year={2024},
      eprint={2401.08001},
      archivePrefix={arXiv},
      primaryClass={cs.NE},
      url={https://arxiv.org/abs/2401.08001}, 
}

@misc{TPINN,
      title={Functional Tensor Decompositions for Physics-Informed Neural Networks}, 
      author={Sai Karthikeya Vemuri and Tim Büchner and Julia Niebling and Joachim Denzler},
      year={2024},
      eprint={2408.13101},
      archivePrefix={arXiv},
      primaryClass={cs.LG},
      url={https://arxiv.org/abs/2408.13101}, 
}

@InProceedings{Glorot_Initial,
  title = 	 {Understanding the difficulty of training deep feedforward neural networks},
  author = 	 {Glorot, Xavier and Bengio, Yoshua},
  booktitle = 	 {Proceedings of the Thirteenth International Conference on Artificial Intelligence and Statistics},
  pages = 	 {249--256},
  year = 	 {2010},
  editor = 	 {Teh, Yee Whye and Titterington, Mike},
  volume = 	 {9},
  series = 	 {Proceedings of Machine Learning Research},
  address = 	 {Chia Laguna Resort, Sardinia, Italy},
  month = 	 {13--15 May},
  publisher =    {PMLR},
  url = 	 {https://proceedings.mlr.press/v9/glorot10a.html}
}

@misc{causal_TT_initial,
      title={Initialization and training of matrix product state probabilistic models}, 
      author={Xun Tang and Yuehaw Khoo and Lexing Ying},
      year={2025},
      eprint={2505.06419},
      archivePrefix={arXiv},
      primaryClass={math.NA},
      url={https://arxiv.org/abs/2505.06419}, 
}

@article{TT_Cross,
title = {TT-cross approximation for multidimensional arrays},
journal = {Linear Algebra and its Applications},
volume = {432},
number = {1},
pages = {70-88},
year = {2010},
issn = {0024-3795},
doi = {https://doi.org/10.1016/j.laa.2009.07.024},
url = {https://www.sciencedirect.com/science/article/pii/S0024379509003747},
author = {Ivan Oseledets and Eugene Tyrtyshnikov}
}

@misc{Unitary,
      title={How to generate random matrices from the classical compact groups}, 
      author={Francesco Mezzadri},
      year={2007},
      eprint={math-ph/0609050},
      archivePrefix={arXiv},
      primaryClass={math-ph},
      url={https://arxiv.org/abs/math-ph/0609050}, 
}

@misc{tn4ml,
      title={tn4ml: Tensor Network Training and Customization for Machine Learning}, 
      author={Ema Puljak and Sergio Sanchez-Ramirez and Sergi Masot-Llima and Jofre Vallès-Muns and Artur Garcia-Saez and Maurizio Pierini},
      year={2025},
      eprint={2502.13090},
      archivePrefix={arXiv},
      primaryClass={cs.LG},
      url={https://arxiv.org/abs/2502.13090}, 
}

@article{Compress_DNN_MPO,
  title = {Compressing deep neural networks by matrix product operators},
  author = {Gao, Ze-Feng and Cheng, Song and He, Rong-Qiang and Xie, Z. Y. and Zhao, Hui-Hai and Lu, Zhong-Yi and Xiang, Tao},
  journal = {Phys. Rev. Res.},
  volume = {2},
  issue = {2},
  pages = {023300},
  numpages = {9},
  year = {2020},
  month = {Jun},
  publisher = {American Physical Society},
  doi = {10.1103/PhysRevResearch.2.023300},
  url = {https://link.aps.org/doi/10.1103/PhysRevResearch.2.023300}
}

\end{document}